\documentclass[runningheads]{llncs}
\usepackage{graphicx}
\begin{document}
\title{FRAUD DETECTION USING DATA-DRIVEN APPROACH}
%
%\titlerunning{Abbreviated paper title}
% If the paper title is too long for the running head, you can set
% an abbreviated paper title here
%
\author{Arianit Mehana\inst{1} \and Krenare Prieva Nuci\inst{2} } 

\institute{TEB SH.A \and
University for Business and Technology, Faculty of Computer Science and Engineering, Kosova \\
\email{am43144@ubt-uni.net, krenare.pireva@ubt-uni.net} }
\maketitle              % typeset the header of the contribution
\begin{abstract}
The extensive use of internet is continuously drifting businesses to incorporate their services in the online environment. One of the first spectrums to embrace this evolution was the banking sector. In fact, the first known online banking service came in 1980. It was deployed from a community bank located in Knoxville, called the United American Bank. Since then, internet banking has been offering ease and efficiency to costumers in completing their daily banking tasks.
The ever increasing use of internet banking and the large number of online transactions, increased fraudulent behaviour also. As if fraud increase wasn’t enough, the massive number of online transactions further increased the data complexity. Modern data sources are not only complex but generated at high speed and in real time as well. This presents a serious problem and a definite reason why more advanced solutions are desired to protect financial service companies and credit card holders.
Therefore, this research paper aims to construct an efficient fraud detection model which is adaptive to costumer behaviour changes and tends to decrease the fraud manipulation, by detecting and filtering fraud in real-time. In order to achieve this aim, a review of various methods is conducted, adding above a personal experience working at a Banking sector, specifically in Fraud Detection office. Unlike the majority of reviewed methods, the proposed model in this research paper is able to detect fraud in the moment of occurrence using an incremental classifier.
The evaluation on synthetic data, based on fraud scenarios selected in collaboration with domain experts that replicate typical, real-world attacks, shows that this approach correctly ranks complex frauds. In particular, our proposal detects fraudulent behaviour and anomalies with up to 97\% detection rate while maintaining a satisfying low cost. 

\end{abstract}
\section{Introduction}
The Internet has been around for decades. Many people have been using it to facilitate their lives and expedite their daily tasks. Of all the aspects of daily life that have benefitted from the internet, the banking sector has been especially effective at capitalizing on internet’s features. It has introduced many attractive ways to increase the scope of its financial services. The emergence of internet banking has allowed banks to offer their customers relatively convenient and flexible banking, also known as e-banking [1]. 
Although there are many advantages of online banking, security issues often discourage customers from using it. This is evolving as many customers have found that the use of online banking could leave their financial assets at risk due to fraudulent activity. 
Fraud is defined as wrongful or criminal deception intended to result in financial or personal gain, or to damage another individual without necessarily leading to direct legal consequences [2]. This ever-growing market urged the need for particular attention to a counter mechanism in order to tone the losses down, which only in the last decade have managed to increase 56,5\% globally [3].
So far, there are different approaches from a number of researchers that in one way or another have proposed techniques to detect these activities, but there is lack of research on detecting these activities in real-time situations. Therefore, this research paper tackles this gap, by proposing a fraud detection approach which uses instance-incremental learning. This methodology increments its knowledge instance by instance which actively and adaptively recognizes such activity in order to prevent it from reaching the final state.
The rest of this research paper is organized as follows: Section 2 presents an analyse on the actual fraud detection mechanisms. Section 3 contains the methodology of the system. The results and the evaluation methods for the proposed model are presented in Section 4. This leads to the last Section, that concludes this research paper and unveils plans for future work.

\section{Related works}
Fraud detection systems come into play when the fraudsters surpass the fraud prevention systems and start a fraudulent transaction. Accordingly, the goal of a fraud detection system is to check every transaction for the possibility of being fraudulent regardless of the prevention mechanisms, and to identify fraudulent ones as quickly as possible after the fraudster has begun to perpetrate. In order to accomplish that, financial institutions use a variety of techniques that can be grouped in three main categories [2]:\\\\
•	Traditional techniques.\\
•	Machine learning techniques.\\
•	Hybrid techniques.\\\\
Besides fraud detecting, the mentioned techniques are widely used in other areas of the banking sector such as [2]: customer retention, marketing, risk management and CRM.\\
\textbf{Traditional techniques} are probably the oldest and most time-proof ones. They consist of defining certain rules and label actions that match them, as anomalous and potentially worth checking. These rules are defined by experts of the financial institutions; therefore, the efficiency of the system depends fully on the them. An example of this kind of approach is presented in [4].Although, it is fast and easy to run as a technique, generating reports every hour makes it hard to detect fraudulent behaviour at the moment of initiation.\\
\textbf{Machine learning techniques} are concerned with general pattern recognition or the construction of universal approximations of relations in the data in situations where no
obvious a priori analytical solution exists [5]. Learning process can be done in a supervised environment or an unsupervised one. 
For supervised learning, best performing methods are: Naïve Bayes, KNN, graphs, artificial neural networks and support vector machines [6]. Some of these methods are presented in [7-10]. Both researchers in [7] and [8] use belief propagation to denote the final score for an instance whether that is a device or an account. This propagation is calculated using the exponentially decaying function. The decaying function is time variant. The higher the time difference between the time of the transaction and the scoring time, the lower the transaction fraud score will be. The model presented in [9] uses neural networks in order to detect fraud on credit card transactions, while in [10], detection is made possible using neural detectors. Neural networks are really complex and require a lot of experimenting in order to produce efficient results. In the other hand, outdated neural detectors in [10] cause a lot of errors in instance classification.\\
Unsupervised learning techniques include: k-means clustering, hierarchical clustering, neural networks, anomaly detection and decision trees. Some examples of unsupervised approaches are presented in [11-13] research work. In the approach presented in [11], fraud is detected by constructing user profiles and excluding deviations using anomaly detection algorithms. Unlike this system, the systems presented in [12] and [13] use k-means clustering in order to group the users. Although, their results are acceptable, detection should be organized so that it compares current transactions with previous  transactions from the same user. Therefore, detection would be more personalized.\\
\textbf{Hybrid techniques} are a combination of two or more computational techniques which provide greater advantages on fraud detecting than individual ones. Most common combinations in hybrid methodologies include [6]:\\\\
•	Traditional techniques with machine learning ones\\
•	Supervised machine learning techniques with unsupervised ones\\\\
An example of each of these categories is presented in [14] and [15], respectively.The proposed approach in [14] combines one-class classification and rule-based methodology at account level, while [15] combines supervised methodologies like decision trees with unsupervised ones like SVM-s. Although the mentioned approaches proudly presented a 97\% fraud detection rate, they lack the instant reaction technology.\\
All previously mentioned techniques in [4,7-15] detect fraud by learning from batches of the dataset. This particular type of operation is called batch learning.
\textbf{Instance-incremental methods} are truly incremental in the sense that they learn from each training example as it arrives. Thus, they can essentially learn indefinitely. This category includes lazy learners and incremental learners such as Naive Bayes Updateable and Hoeffding Trees [16].

\section{Design and Implementation of proposed model}
The presented model is built purposely on a synthetic dataset, keeping confidentiality intact. This research paper uses a dataset [17] as secondary data for simulation purposes. This dataset contains mobile transactions based on a sample of real transactions extracted from one month of financial logs from a mobile transaction service. The original logs were provided by a multinational company, who is the provider of the mobile financial service which is currently running in more than 14 countries all around the world. Additionally, in order to evaluate the selected dataset and also the proposed model, a primary data collection was conducted using interview methods. The interviews were organised with domain experts from local banks. Initially, this dataset was pre-processed with a number of filters and scaled down to a smaller and more balanced version from the initial one. Pre-processing the data helped in creating a better understanding for the upcoming analysis and structuring of the model, that was done using WEKA software. WEKA is an open source machine learning software that can be accessed through a graphical user interface, standard terminal applications, or a Java API. It is widely used for teaching, research, and industrial applications, contains a plethora of built-in tools for standard machine learning tasks, and additionally gives transparent access to well-known toolboxes such as scikit-learn, R, and Deeplearning4j.  Additionally, WEKA offers a number of incremental classifiers including: Hoefding tree, Knn lazy learner and Naïve Bayes Updateable. This model is build using the last mentioned classifier while Hoefding tree and KNN were also separately tested in order to compare the results. Naïve Bayes Updatable is the successor of Naïve Bayes, that can handle evolving data, which is always the case with any transaction dataset as in [17]. It is efficient and requires less memory usage compared to its predecessor. For better segregation with other approaches, this model will be named as “Active Fraud Detection Model”, hereafter mentioned as AFDM. For testing purposes, this model was validated using k-fold cross-validation, which is a resampling procedure used to evaluate machine learning models. This procedure has a single parameter called k that refers to the number of groups that a given data sample is to be split into. These k groups are then recursively used for testing and training. The results of the tests are evaluated using the confusion matrix and cost. 
%kjo figure nuk po del ne kete vend por tek kapituli tjeter!
\begin{figure}
\begin{center}
\includegraphics[width=90mm]{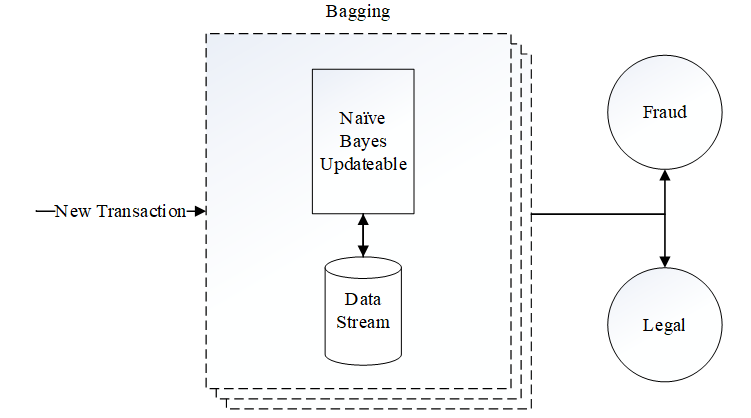}
\caption{Architecture of AFDM.} \label{fig1}
\end{center}
\end{figure}

As shown in Figure 1, AFDM not only classifies instances in the moment of occurrence, it also uses each instance to immediately train the learner. Any time a new transaction is made, this model classifies it either as fraud or as legal. Afterwards it uses that knowledge and updates Naïve Bayes in order to make it effective for similar upcoming instances. Each step is enhanced using the bagging technique. It basically generates a number of new training sets in a number of iterations and outputs the aggregated average which also reduces variance and helps to avoid overfitting.

\section{Evaluation of proposed model}
AFDM was evaluated against a couple of other batch and incremental classifiers in the same conditions. The flow and the results of this test process are unveiled in this section.The concurrent results urged the need to use an additional performance measure, called cost. Cost is a value calculated from the sum of misclassified instances in both classes, multiplied by their relative weight.

\begin{figure}
\begin{center}
\includegraphics[width=90mm]{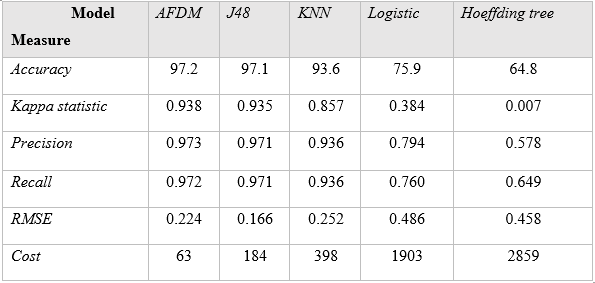}
\caption{Summary of test results.} \label{fig2}
\end{center}
\end{figure}

Table 1 expresses the tests results using the mentioned classifiers. Although, AFD model performed better than other classifiers, J48 presented abutting results. Actually, RMSE value of J48 is lower than that of the presented model, but it’s the opposite with the cost measure. This is mainly because J48 did a better classifying on the legal class but missed a lot of fraud instances, which affect the cost more, based on their higher weight. KNN also managed to output a pleasing accuracy, while Logistic and Hoeffding Tree conveyed a poor performance compared to the AFDM. Putting AFDM aside, batch learners seem to do a better classifying compared with incremental ones, lacking the real-time prediction.

\section{Conclusion}
This research paper introduced a novel approach for real-time fraud detection in online banking transactions using incremental learning approach. AFDM is a supervised fraud detection model, built to aid the fraud unit in a financial institution. Its findings contributed to domain experts achieve a better understanding of the potential risk they are exposed to. The development of this model was made possible by gratifying the objectives precisely. Initially, the most successful fraud detection methodologies were studied. The disadvantages that surfaced from these models built problems that needed to be solved by the proposed AFDM. 
Unlike other approaches, this model is transaction-based. Analysing transactions instead of accounts not only allowed a more detailed dissection, it also served a better environment for post detection actions. Additionally, AFDM accounted all of the user transactions which is a must. This excluded the need of a time window and reduced the risk of missing a fraudulent behaviour. On top of that, a classifying algorithm like Naïve Bayes Updateable incremented its knowledge transaction by transaction. This learning methodology provided the ability to detect and respond in real time. It also allowed to learn new concepts of behaviour changes immediately. 
The effectiveness of this model was evaluated on a dataset modelled from a mobile transaction service. The resemblance with a real dataset offered real-world scenarios and ensured valid results. Comparing these results with classifiers from different categories ultimately proved the significance of this model. 
Given the good results and the consistency presented in the previous section, AFDM is undoubtedly an attractive pick for the services it offers in the fraud detection domain. Currently this model presents the framework for fraud detection classification. Among the directions for future work it is planned to implement global-based classifying and post detection actions. Global-based classifying will deal with new costumers that have a small number of transactions. On the other hand, post detection actions will enable the interaction with costumers. In case of a suspicious behaviour, the transaction will be blocked until the costumer proves the opposite.

%
% ---- Bibliography ----
%
% BibTeX users should specify bibliography style 'splncs04'.
% References will then be sorted and formatted in the correct style.
%
% \bibliographystyle{splncs04}
% \bibliography{mybibliography}
%

\end{document}